\pdfoutput=1

\documentclass[11pt]{article}
\usepackage{xcolor}

\usepackage{EMNLP2023}

\usepackage{times}
\usepackage{latexsym}
\usepackage{graphicx}
\usepackage{listings}
\usepackage{booktabs}
\usepackage{hyperref}
\usepackage[T1]{fontenc}

\usepackage[utf8]{inputenc}

\usepackage{microtype}

\usepackage{inconsolata}

\usepackage{enumitem}

\title{From Nuisance to News Sense: \\Augmenting the
News with Cross-document Evidence and Context}

\author{Jeremiah Milbauer \and Ziqi Ding \and Zhijin Wu \and Tongshuang Wu\\
\\
Carnegie Mellon University, Pittsburgh PA, USA\\
\texttt{ \{jmilbaue | sherryw\}@cs.cmu.edu} \\
\texttt{ \{ziqiding | zhijinw\}@andrew.cmu.edu}
}

\begin{document}
\maketitle
\begin{abstract}
Reading and understanding the stories in the news is increasingly difficult. Reporting on stories evolves rapidly, politicized news venues offer different perspectives (and sometimes different facts), and misinformation is rampant. However, Existing solutions merely \textbf{aggregate} an overwhelming amount of information from heterogenous sources, such as different news outlets, social media, and news bias rating agencies.
We present \textsc{NewsSense}, a novel sensemaking tool and reading interface designed to collect and \textbf{integrate} information from multiple news articles on a central topic.
\textsc{NewsSense} augments a central, grounding article of the user's choice by linking it to related articles from different sources, providing inline highlights on how specific claims in the chosen article are either supported or contradicted by information from other articles. Using \textsc{NewsSense}, users can seamlessly digest and cross-check multiple information sources without disturbing their natural reading flow.
Our pilot study shows that NewsSense has the potential to help users identify key information, verify the credibility of news articles, and explore different perspectives. We opensource NewsSense at \href{https://www.github.com/jmilbauer/NewsSense}{github.com/jmilbauer/NewsSense}, and a demo video is hosted at \href{https://youtu.be/2D5LYbsQJak}{youtu.be/2D5LYbsQJak}.

\end{abstract} %
\section{Introduction}

\begin{figure}[t!]
\centering
    \includegraphics[width=\linewidth]{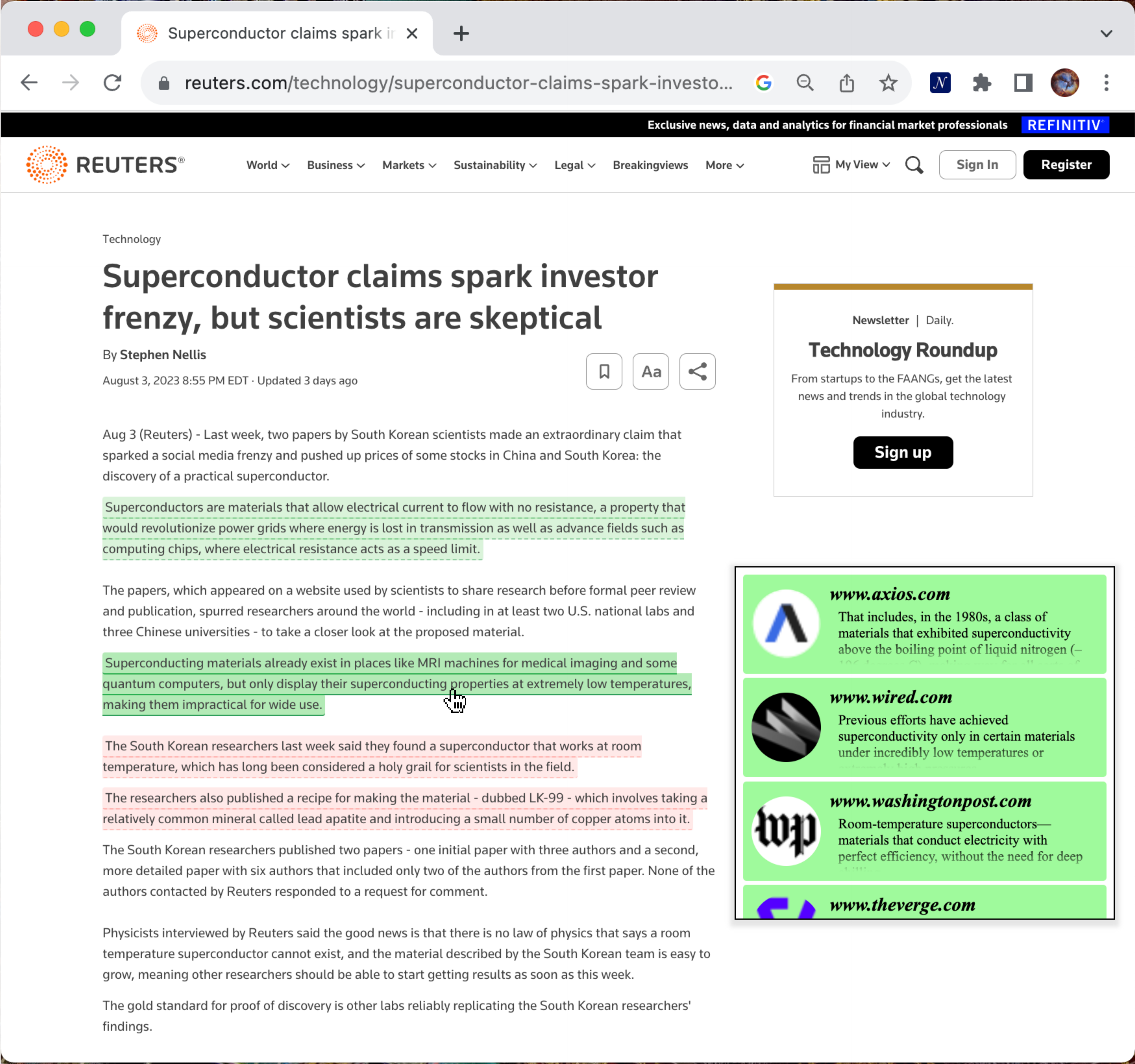}
    \caption{A screenshot from \textsc{NewsSense} browser extension running in Chrome. The extension provides highlights indicated supported and controversial information. When the user clicks on a highlighted sentence, \textsc{NewSense} adds an scrollable overlay box containing snippets of external evidence.}
    \label{fig:interface_overview}
    \vspace{-15pt}
\end{figure}

Why is it so hard, and so exhausting, to read the news? In the quest for knowledge, news readers today must contend with a rapidly evolving 24-hour news cycle, multiple news venues competing for attention and clicks, and the challenge of integrating fact-based reporting, opinion pieces, and social media commentary \cite{zittrain, benkler, posttruth}. With news becoming increasingly politicized \cite{faris} readers also face the challenge of identifying and avoiding misinformation, disinformation, and hyperbolic ``clickbait" as they try to remain informed about the world around them.

Various solutions have been proposed to assist with users' news reading. 
For example, media watchdog companies have created media bias charts to represent political leaning and credibility of news sources \footnote{https://www.allsides.com/media-bias/media-bias-rating-
methods} \footnote{https://adfontesmedia.com/interactive-media-bias-chart/}. However, these resources force users to rely on potentially untrustworthy third-party designations of media bias, which treat each \emph{news source} as a whole, without digging into specific articles or topics.

While novel automatic fact checking~\cite{fever} and fake news detection~\cite{fakesurvey, clickbaitdetection} systems can provide verification per-article, these approaches typically rely on a preordained corpus of verified facts which cannot keep up with the always-evolving facts, and may not reflect user preferences or multilateral perspectives. 
Aggregating articles from heterogeneous sources seem a more promising direction for cross-checking new facts (without predetermined groundtruths) and collecting different perspectives, but existing attempts are still too coarse and overwhelming. 
For example, both Google News' ``Stories" \footnote{https://news.google.com/stories/} feature and Ground.news\footnote{https://ground.news} collect articles about the same events but display them in the form of  exhaustive lists -- Users are still forced to read and compare each article on its own.

We argue that instead of simply collecting and aggregating news articles, information and claims from multiple sources should be \textbf{integrated} in a way that allows users to identify fine-grained claim-level bias, spin, controversy, or evidence.

We present \textsc{NewsSense}, a novel framework for sensemaking within a cluster of documents, to address the three key problems of news reading -- bias, factuality, and article overload -- in a single streamlined interface.
\textsc{NewsSense} leverages existing modular NLP techniques to identify and link claims made across \emph{a cluster of news articles}, such that these articles become references for each other. 
\textsc{NewsSense} also displays the linking information using an \emph{interactive reading interface}, which allows users to easily explore the cross-document connections without being overwhelmed. 
In pilot user studies, we see that the \textsc{NewsSense} framework has the potential to help users identify key information, verify the credibility of news articles, and explore different perspectives.

While \textsc{NewsSense} is primarily implemented for news articles, our framework can be easily generalized to other assisted reading and cross-checking scenarios (e.g., compare multiple manuscripts in literature reviews). The key contributions of \textsc{NewsSense} are:
\begin{enumerate}[nosep, leftmargin=1.4em,labelwidth=*,align=left]
    \item A pipeline for analyzing the connections between a collection of documents.
    \item A two-stage method for efficiently computing cross-document links between claims that support or contradict each other, enabling ``reference-free'' fact checking.
    \item A framework for visualizing cross-document connections, and integrating claims from multiple documents into a single reading experience.
\end{enumerate}
We conclude by discussing the generality and potential social benefits of \textsc{NewsSense}.

\section{Related Work}
This section covers related research across media analytics, sensemaking, and natural language processing. Though some core ideas of this work have been explored in the past, to our knowledge they have never been combined in a single system.

\paragraph{Media Bias and Analytics}
Research on media bias includes academic research to study social media sharing patterns \cite{mediacloud2021, bakshy2015exposure} and bias within media publications \cite{flaxman2016filter, hamborg2019automated, groseclose2005measure}. Commercial products exist in this area as well, such as the media bias charts of AllSides \footnote{https://www.allsides.com/media-bias/media-bias-rating-methods}, which classifies political slant into one of five categories, and Ad Fontes Media \footnote{https://adfontesmedia.com/interactive-media-bias-chart/} , which models both political slant and factual credibility.

Research on news and social content aggregation has focused primarily on headline detection, timeline construction and clustering \cite{newsclustering, laban2017newslens}, and event detection \cite{twitterevent, newsevent}. There exist user-oriented products in this space, such as Google News Stories \footnote{https://news.google.com}, and Ground.news \footnote{https://ground.news/}. Some outlets, such as Propublica, aggregate their news stories into timelines \footnote{https://www.propublica.org/series}.

\paragraph{Reading Interfaces and Sensemaking}
Recent work on reading interfaces has primarily focused on scientific literature, augmenting documents with information about cited papers \cite{semanticreader, threddy}, or augmenting references within the documents themselves \cite{scholarphi}.

For the News domain specifically, \citet{laban2017newslens} aggregates articles and extracts key quotes to construct a timeline for a given story. We are also aware of an abstract describing work to combine multiple article headlines and ledes into a single digestible form, though no follow-up is available  \cite{glassmanTriangulating}.

\paragraph{Fact Verification and NLI}

Natural Language Inference is a task focused on classifying the relationship between a pair of sentences as either ``neutral", ``entailment", or ``contradiction." Datasets such as SNLI \cite{snli} and MNLI \cite{mnli} have become major benchmarks for natural language processing research. Recent work has also considered document-level NLI \cite{contractnli, propsegment}, as well as cross-document reasoning based in NLI \cite{sentnli}, and scalable pairwise reasoning \cite{lait}.

There is also a growing body of work on NLP systems for fact verification and attribution. Recent datasets include FEVER \cite{fever} and VitaminC \cite{vitaminc}, as well as datasets focused on real-world examples of updating, editing, and citing claims in domains like news and Wikipedia \cite{wikiverify, newsedits, fruit}. %
\section{The \textsc{NewsSense} Framework}
\label{sec:newsreader-framework}

The core philosophy behind \textsc{NewsSense} is to go beyond article aggregation by integrating the information contained within a cluster of related news articles into the reading experience.

\textsc{NewsSense} starts with a single ``focus" article and a set of related ``background" articles. The distinction between focus and background article is arbitrary, as any article within the cluster could be designated the focus article. \textsc{NewsSense} then identifies claims within the focus article that are related -- either by contradiction or entailment -- to claims within the background articles. The claims in the focus article are then highlighted, and linked to the background articles so that users can explore the supporting or contradicting evidence for a given claim without just relying on third-party measurements of bias or credibility.

The \textsc{NewsSense} interface has three primary components: a Focus Article, Sentence Highlights, and External Evidence. Together, these elements display the computed connections between the focus article and the background articles.

\subsection{Focus Article}

The \textsc{NewsSense} interface features a central panel that displays the focus article, including the entire news article the reader is interested in primarily. The focus article can be presented through a dedicated applciation, or by adding \textsc{NewsReader} as an overlay on top of the existing web browsing experience.

\subsection{Sentece Highlights}

The interface highlights claims made in the focus article that have supporting articles in green underlines, while claims with contradicting articles are highlighted in red underlines. By doing so, readers can quickly and easily identify areas of agreement and disagreement across different news sources.

\subsection{External Evidence}

When the reader hovers or clicks on a highlighted claim, an overlay panel appears, containing the supporting or contradicting claim excerpts, as well as their news sources. Links can be annotated with information about the general credibility or political slant of the referenced news venue.

\begin{figure*}[t!]
    \centering
    \includegraphics[width=\textwidth]{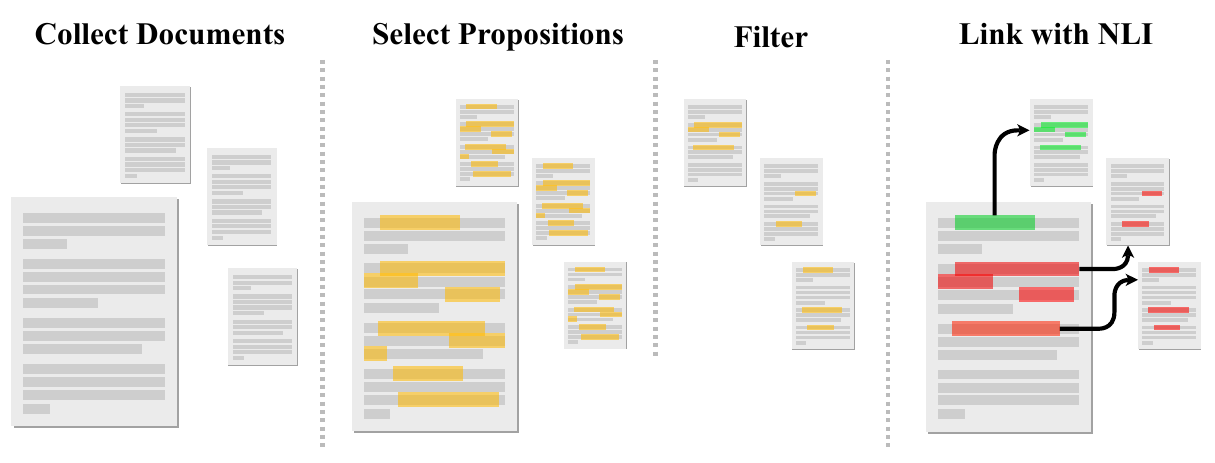}
    \caption{The four stages of the \textit{NewsReader} linking pipeline: Article collection, Claim detection, claim filtering, and claim linking}
    \label{fig:pipeline_diagram}
\end{figure*}

For readers’ convenience, each supporting or contradicting claim is clickable and directs the reader to origin of the associated claim. This allows readers to quickly access the relevant claims without having to search through an entire article. For a standalone \textsc{NewsSense} interface, readers will be prompted with a ``back” button in the secondary articles to quickly go back to the focus article. By providing this functionality, readers can easily navigate through the focus and secondary articles and compare viewpoints, further enhancing their understanding of the news story. %
\section{Pilot Study}

To gather feedback on the proposed \textsc{NewsSense} interface and provide insights for the actual implementation, we conducted a pilot user study using a NewsReader mockup built with Figma \footnote{https://www.figma.com/}. This section describes the design and results of the study.

\subsection{Study Design}

We aim to collect feedback on \textsc{NewsSense}'s basic functionality, interface design, and content quality.

The participants were assigned a task of reading a news article using \textsc{NewsSense} and answering a set of questions. The questions focused on the content of the news, how and where the user located information, and their level of trust in the information. These questions aimed to assess the basic functionality of \textsc{NewsSense} in helping readers understand news comprehensively, to motivate further development of the system.

\begin{figure}[t!]
\centering
    \includegraphics[width=\linewidth]{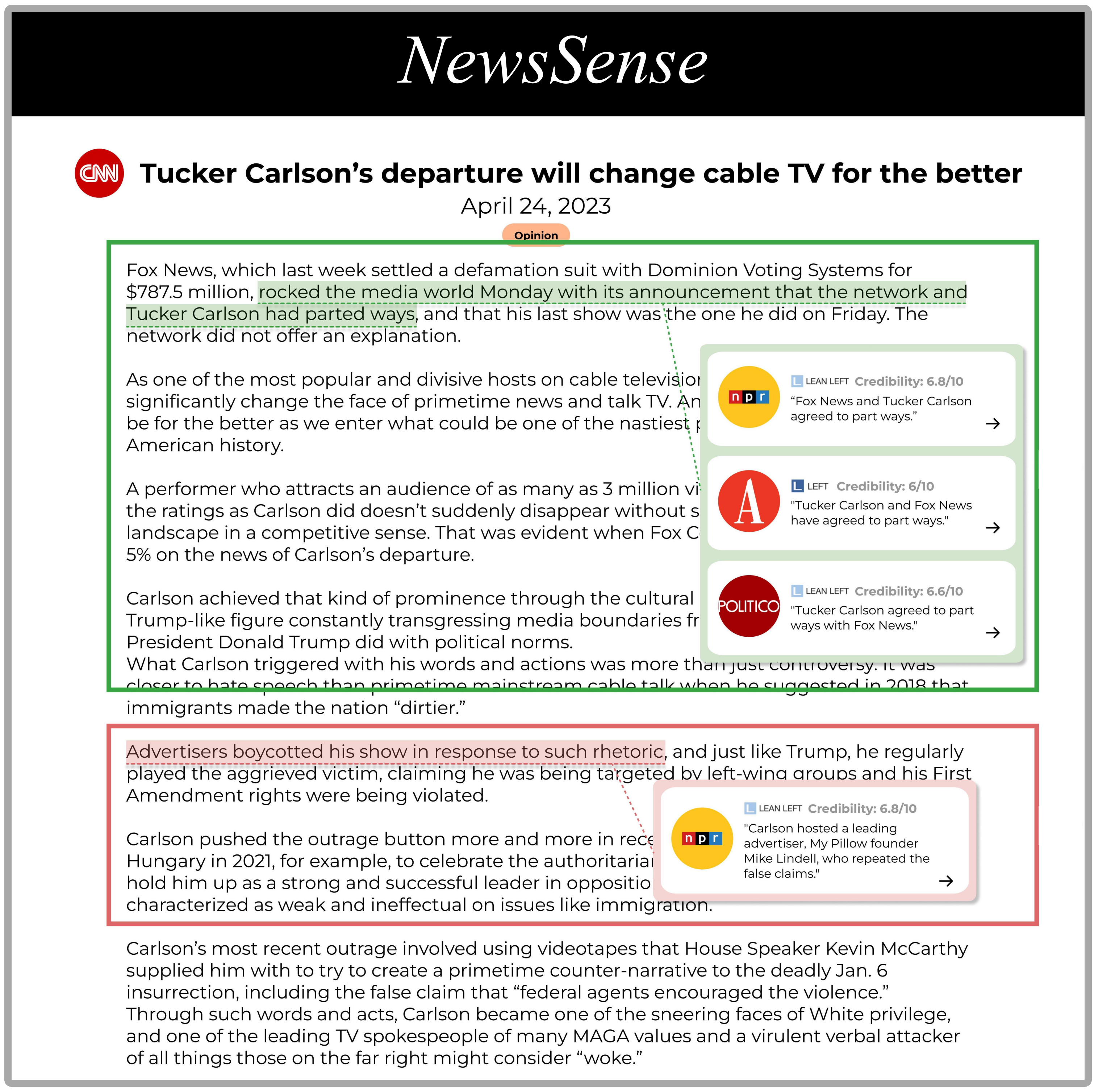}
    \caption{The design layout of the pilot study, prototyped in Figma. The article is presented in a central panel, featuring claims with supporting articles highlighted in green (section boxed in green), and claims with contradicting articles highlighted in red (section boxed in red). Each claim has an associated overlay box of external evidence that appears when the user hovers over the text.}
    \label{fig:user_experience}
    \vspace{-15pt}
\end{figure}

\subsection{Results}

Following the pilot user study with over 10 users, we identified several key findings. First, all users found \textsc{NewsSense} to be useful in locating important information and verifying the credibility of news articles, aligning with our initial goal. The user-friendly interface of \textsc{NewsSense} was well-received, though participants suggested enhancing interactivity to set it apart from other solutions. For instance, displaying real-time feedback like \textit{``NewsSense is analyzing the article"} during loading.

Regarding content quality, some users found \textsc{NewsSense} limited and suggested increased labeling or categorization within articles. One user noted \textit{Two highlighted sentences per page are insufficient for in-depth analysis."} User preferences varied for article summarization, with some wanting more key points and others preferring brevity. Contradicting previous feedback, one user preferred \textit{``Summarizing key points only, rather than selecting sentences with unclear relevance."} Addressing this, \textsc{NewsSense} could allow customization, letting users choose key point count and filter supporting/contradicting data.

\subsection{Study Takeaways}

We found that users liked how \textit{NewsSense} highlighted important sentences from an article. We realized that the claims which are consistent across multiple articles (ie, those which are supported at least once) are likely to be the most important aspects to a given story. \textsc{NewsSense} could inform readers when there are key claims from across the article cluster missing from the article they are reading.

We also found that the bias labels for news venues could be overwhelming, and including them ran counter to our aim of reference-free verification; we eliminated these labels.

Users also appreciated how highlighted sentences functioned as summaries. Consequently we enhance the visibility of text highlights and further emphasize the alignment or contradiction of specific source by making the External Evidence cards colored accordingly. %
\section{System Overview}

Following our user study, we implemented the \textsc{NewsSense} framework as a browser plugin, which adds augmentations to news articles encountered on the web. Figure \ref{fig:interface_overview} shows the final appearance of the browser plugin. Code for system and plugin can be accessed at \href{https://www.github.com/jmilbauer/NewsSense}{github.com/jmilbauer/NewsSense}, and a demo video can be viewed at \href{https://youtu.be/2D5LYbsQJak}{youtu.be/2D5LYbsQJak}.

This section provides a description of the natural language processing system which powers \textsc{NewsSense}. Figure \ref{fig:pipeline_diagram} illustrates the four general steps of the pipeline: Collection, Selection, Filtering, and Linking.

\subsection{Article Collection}

First, we must collect a cluster of news articles that are all about the same news event. Our implementation scrapes data from Google News Stories, a website that collects many articles about the same events across news venues. After collecting article URLs via Google News Stories, we then collect the content of each article. A typical story contains over 50 articles.

\subsection{Claim Selection}

The next phase of the pipeline is to select the claims within each article cluster. We initially assumed a 1-to-1 mapping between sentences and claims, but quickly found that news articles often contain complex multi-clause sentences, which are not suitable for natural language inference. To address this issue, we few-shot prompting to generate a list of claims from sentences using a large language model (LLM). In our experiments, prompt exemplars are drawn from the \textsc{PropSegmEnt} dataset \cite{propsegment}, and the LLM used is OpenAI \texttt{text-davinci-003}. Full prompt details are provided in Appendix \ref{sec:claim-prompt}. We also note that the authors of \textsc{PropSegmEnt} report that \texttt{T5-Large} performs reasonably well on the task, suggesting the possibility for further pipeline improvements.

\subsection{Claim Filtering}

Articles often contain over 30 sentences. For a cluster of 50 articles, a pairwise comparison of the full cartesian product of sentences has $\mathcal{O}((NL)^2)$, which is in practice well over 1,000,000 comparisons. Performing this level of computation at scale, even if we are pre-computing results for each article cluster, is simply not feasible. To address this, we perform an initial filtering step with leverages the fact that the vast majority of claims across any two articles are unrelated. We consider two approaches for claim filtering: Embedding Similarity filtering (ES) and Lexical Overlap filtering (LeO).

For Embedding Similarity filtering, we encode each claim in each article using a Transformer-based sentence encoder. Then, for each claim we retain only the $k$ most similar other claims for comparison. In our implementation, we use the Sentence Transformers \cite{sentencetransformers} \texttt{all-Mini-LM-L6-v2}.

For Lexical Overlap filtering, we compare each sentence only with sentences that have overlapping words, as these sentences are likely to discuss similar topics. In our implementation, we process claims by first remove stopwords, then stemming using the NLTK \cite{nltk} implementation of the Porter Stemmer \cite{porterstemmer}, and compute overlap scores using the Jaccard Index.

We evaluated each filtering method on the MNLI \cite{mnli} validation data, treating pairs of randomly sampled sentences as negative examples, and labeled ``entailment" and ``contradiction" sentence pairs as positive examples. For ES, we set a threshold of $0.3$ cosine similarity; for LeO, we set a threshold of $0.1$ overlap. We note that \texttt{all-Mini-LM-L6-v2} included MNLI in its training data.

\begin{table}[t]
\resizebox{\linewidth}{!}{
    \begin{tabular}{lllll}
    \toprule
                         & Prec.   & Rec.   & Macro-F1 & TNR \\ \midrule
    Embedding Similarity & 0.9905  & 0.9528 & 0.9718   & 0.9909 \\
    Entity Overlap       & 0.9116  & 0.8891 & 0.9014   & 0.9138  \\
    \bottomrule
    \end{tabular}
}
\vspace{10pt}
\caption{The positive-class precision, recall, macro-averaged F1, and true negative rate for the two filtering methods. Embedding Similarity outperforms entity overlap on every metric.}
\label{tab:filter_table}
\vspace{-10pt}
\end{table}

We include a summary of the results of these experiments in Table \ref{tab:filter_table}, which indicates that the ES method outperforms the LeO method. Of particular interest is the true negative rate, as this indicates the percentage of non-related sentences we expect to filter out.

\subsection{Claim Linking}

Once claim pairs have been filtered, we classify each pair according to the Natural Language Inference (NLI) framework, as ``entailment," ``contradiction," or ``neutral." We employ a pretrained language model, \texttt{RoBERTa} \cite{roberta}, which was then fine-tuned on MNLI \cite{mnli}, a popular dataset for NLI. We download this fine-tuned version of \texttt{RoBERTa} from the Hugging Face model library \footnote{https://huggingface.co/roberta-large-mnli}. To avoid clutter, we keep fewer than 100 of the most confident predictions for each positive class (entailment or contradiction) within the article cluster. Claims are then assigned back to the sentences from which they were generated, and the sentence pairs are linked. %
\section{Discussion}

\textsc{NewsSense} provides an intuitive and effective interface for integrating information from a large cluster of news articles into a single, focused reading experience. Although applied in this demo to news articles, the \textsc{NewsSense} framework could just as easily be applied to the analysis of other types of document clusters as well. The pipeline itself is highly modular, and can easily adopt advancements in NLP technologies to increase the accuracy or decrease processing time.

\subsection{Future Work}
The generality of the \textsc{NewsSense} also introduces a number of opportunities for future development.
\paragraph{Expanding the Scope of NewsReader}
Often, articles contain references to past events. In the future, we would like to explore the possibility of extending the \textsc{NewsSense} framework beyond the immediately article clusters to include all relevant articles in a timeline of events.

Additionally, as we explored the \textsc{NewsSense} framework, we noticed that the clustering approach we used -- the Google News Stories -- sometimes established associations between source \textit{news} articles, and background \textit{primary source} articles. As a result, we would encourage further exploration of the \textit{NewsSense} framework applied to heterogenous and primary-source document collections, whcih might include primary scholarly literature.

\paragraph{NLP Pipeline Improvements} Because the \textsc{NewsSense} pipeline is modular, a number of improvements can be explored. For sentence segmentation, methods that use a fine-tuned language model could improve the speed of segmentation. For sentence filtering, we relied on a pretrained sentence retrieval model -- but future iterations of \textsc{NewsSense} could use a sentence retrieval model fine-tuned on ``unrelated" pairs of MNLI sentences.
We also found that in many cases, the NLI algorithm used for claim linking made mistakes, perhaps owing to the fact that news articles may not perfectly match models trained on MNLI. Other NLI approaches could be explored, such as SeNtLI \cite{sentnli} (which is designed to work for both individual premises and longer sentences) and LAIT \cite{lait}, which speeds up inference time through late interaction.

\paragraph{More Useful Information} Our final version of \textsc{NewsSense} focused on a relatively paired-down and streamlined interface. However, users did suggest that they would like to see article summaries, and we identified that in many cases key information is repeated across multiple articles. We would consider adding a way for \textsc{NewsSense} to convey the highlights -- key claims from across the article cluster -- when a user is reading an article. We noticed other forms of unintended but incredibly useful functionality: For example, as stories develop, new facts emerge that may contradict old ones. This means that newer articles might supersede older ones. Future iterations of \textsc{NewsSense} should help readers understand when a contradiction may be due to evolving stories .

\paragraph{Deployment} A larger-scale user study would help determine what further improvements could be made to the framework. Our fully interactive interface would help us run a study at larger scale. %
\section{Conclusion}
We presented a novel framework for sensemaking within a cluster of documents. We applied this framework to news articles, building \textsc{NewsSense}, an interactive tool that links claims within one document to supporting or contradicting evidence across the entire document cluster. \textsc{NewsSense} assists readers by helping them to understand the connections and perspectives across many documents. Readers can thus attain a more comprehensive understanding of a given subject, while avoiding the dangers of information overload. Crucially, \textsc{NewsSense} provides a framework for \textit{reference-free} fact verification, which is essential in domains such as the news where events evolve in real time, because a knowledge source for factual grounding may not be available.

Our work expands the growing body of literature on natural language processing applications to document-level sensemaking by demonstrating the utility of automatically generated cross-document links, as well as the application of sensemaking tools to the news reading experience.
\section*{Limitations and Ethics}

\textit{NewsSense} falls within the genre of computer science literature that aims to solve problems such as misinformation. A broad critique of this literature is that it falls within the realm of \textit{techno-solutionism}, in the sense that we seek to develop technological solutions to problems that are potentially social in origin, and perhaps better solved with a socially-oriented approach.

However, we posit that because the problem of misinformation propagation and newsmedia overload are both enabled by technology, we do have a responsibility to explore the ability of technological systems to address these challenges. Unlike approaches that involve traditional fact verification, the reference-free approach of \textit{NewsSense} does not take on the role of deciding what is true and what is not; it simply helps users understand the context of each claim, and make their own decisions.

Beyond this critique, we have also understand there are potential obstacles to the use of a system like \textit{NewsSense}. The people who choose to use a system such as \textit{NewsSense} may already be predisposed to consider and critically evaluate diverse perspectives in the news; \textit{NewsSense} may not be adopted by who needs it most. We also consider that the highlighted links may clutter the reading experience, but we believe this concern is mitigated by the fact that news websites are already quite cluttered (by ads, sponsored links, and article thumbnails) and that users found the highlights helpful in identifying the key components of the articles. %

\bibliography{anthology, custom}
\bibliographystyle{acl_natbib}

\appendix
\section{Appendix: Prompt for Claim Extraction}
\label{sec:claim-prompt}

This is the prompt used for claim extraction from news article sentences:

\begin{lstlisting}
Extract all the claims from a sentence, ignoring extraneous words such as unimportant adverbs. A sentence may contain multiple claims. Each claim should be of the form <subject> <predicate> <object>, and should have the first occurrence of any pronouns replaced by their antecedents.

Sentence: "The 3rd and 4th stations all announced that they would be postponed, and the Monaco station was subsequently cancelled."
Claim: Monaco station was cancelled.
Claim: 4th stations announced they would be postponed.
Claim: The 3rd stations announced they would be postponed.
Claim: The 4th stations postponed.
Claim: The 3rd stations postponed.

Sentence: "Lewis Hamilton and Mercedes have once again confirmed themselves as drivers and constructors world champions."
Claim: Mercedes confirmed themselves as constructors world champions.
Claim: Lewis Hamilton confirmed themselves as drivers world champions.

Sentence: "Local organizers in East Palestine, Ohio on Monday said their activism has successfully pressured rail company Norfolk Southern to agree to a limited relocation plan for some residents affected by last month's train derailment, but added they have no intention of backing down from their demand for justice for thousands of people in the area who are struggling in the aftermath of the accident."
Claim: Local organizers said their activism has pressured rail company Norfolk Southern to agree to a limited relocation plan.
Claim: Local organizers have no intention of backing down from their demand for justice.
Claim: Rail company Norfolk Southern agree to a limited relocation plan.

Sentence: <INSERT SENTENCE HERE>
\end{lstlisting}

\end{document}